\documentclass{egpubl}
\usepackage{pg2023}

\SpecialIssuePaper         

\CGFStandardLicense

\usepackage[T1]{fontenc}
\usepackage{dfadobe}

\usepackage{cite}  
\BibtexOrBiblatex
\electronicVersion
\PrintedOrElectronic
\ifpdf \usepackage[pdftex]{graphicx} \pdfcompresslevel=9
\else \usepackage[dvips]{graphicx} \fi

\usepackage{egweblnk}

\usepackage{times}
\usepackage{epsfig}

\usepackage{graphicx}
\usepackage{amsmath}
\usepackage{amssymb}
\usepackage{booktabs}
\usepackage[utf8]{inputenc} 
\usepackage[T1]{fontenc}    
\usepackage{url}            
\usepackage{booktabs}       
\usepackage{amsfonts}       
\usepackage{nicefrac}       
\usepackage{microtype}      
\usepackage{xcolor}         
\usepackage{graphicx}
\usepackage{amsmath}
\usepackage{amssymb}
\usepackage{makecell}
\usepackage{engord}
\usepackage{adjustbox}
\usepackage{enumitem}
\usepackage{ifthen}

\usepackage{caption}

\usepackage{xspace}
\usepackage{soul}

\usepackage[capitalize]{cleveref}
\crefname{section}{Sec.}{Secs.}
\Crefname{section}{Section}{Sections}
\Crefname{table}{Table}{Tables}
\crefname{table}{Tab.}{Tabs.}

\newcommand{\shortname}{MMFS\xspace}

\newcommand{\loss}{\mathcal{L}}
\newcommand{\lossfirststage}{\loss_{\mathrm{stage-1}}}
\newcommand{\losssecondstage}{\loss_{\mathrm{stage-2}}}
\newcommand{\losszeroshot}{\loss_{\mathrm{0-shot}}}
\newcommand{\lossoneshot}{\loss_{\mathrm{1-shot}}}

\newcommand{\encoder}{\mathbf{E}}
\newcommand{\decoder}{\mathbf{D}}
\newcommand{\generator}{\mathcal{G}}
\newcommand{\discriminator}{\mathcal{D}}

\newcommand{\clipdirection}{\mathbf{d}_{\mathrm{CLIP}}}
\newcommand{\clipfeature}{\mathbf{F}_{\mathrm{CLIP}}}
\newcommand{\cliplayertoken}{T^l_{\mathrm{CLIP}}}

\newcommand{\positionalembedding}{\mathbf{e}_{\mathrm{pos}}}

\newcommand{\mappingnetwork}{\mathbf{M}}

\newcommand{\prompt}{p}
\newcommand{\textprompt}{\prompt_{\mathrm{text}}}
\newcommand{\imageprompt}{\prompt_{\mathrm{image}}}

\newcommand{\referimage}{\mathbf{I}_{ref}}
\newcommand{\reconimage}{\mathbf{I}_{rec}}
\newcommand{\styleimage}{\mathbf{I}_s}
\newcommand{\inputimage}{\mathbf{I}_r}

\newcommand{\citep}[1]{\cite{#1}}

\newcommand{\revision}[1]{#1}
\newcommand{\delete}[1]{}

\renewcommand{\paragraph}[1]{\noindent \textbf{#1}}

\title[Multi-Modal Face Stylization with a Generative Prior]{Multi-Modal Face Stylization with a Generative Prior}

\author[Li et al.]
{\parbox{\textwidth}{\centering Mengtian Li$^{1*}$\orcid{0000-0001-6724-6177},
      \hspace{2pt}
      Yi Dong$^{2*}$\orcid{0009-0008-8880-0606},
      \hspace{2pt}
      Minxuan Lin$^{1}$\orcid{0009-0006-5130-5754},
      \hspace{2pt}
      Haibin Huang$^{1\dagger}$\orcid{0000-0002-7787-6428},
      \hspace{2pt}
      Pengfei Wan$^{1}$\orcid{0000-0001-7225-565X},
      \hspace{2pt}
      and Chongyang Ma$^{1}$\orcid{0000-0002-8243-9513}
        }
        \\
{\parbox{\textwidth}{\centering $^1$Kuaishou Technology, China \hspace{0.3in}
        $^2$Tsinghua University, China
      }
}
}

\begin{document}

\teaser{
 \includegraphics[width=0.76\linewidth]{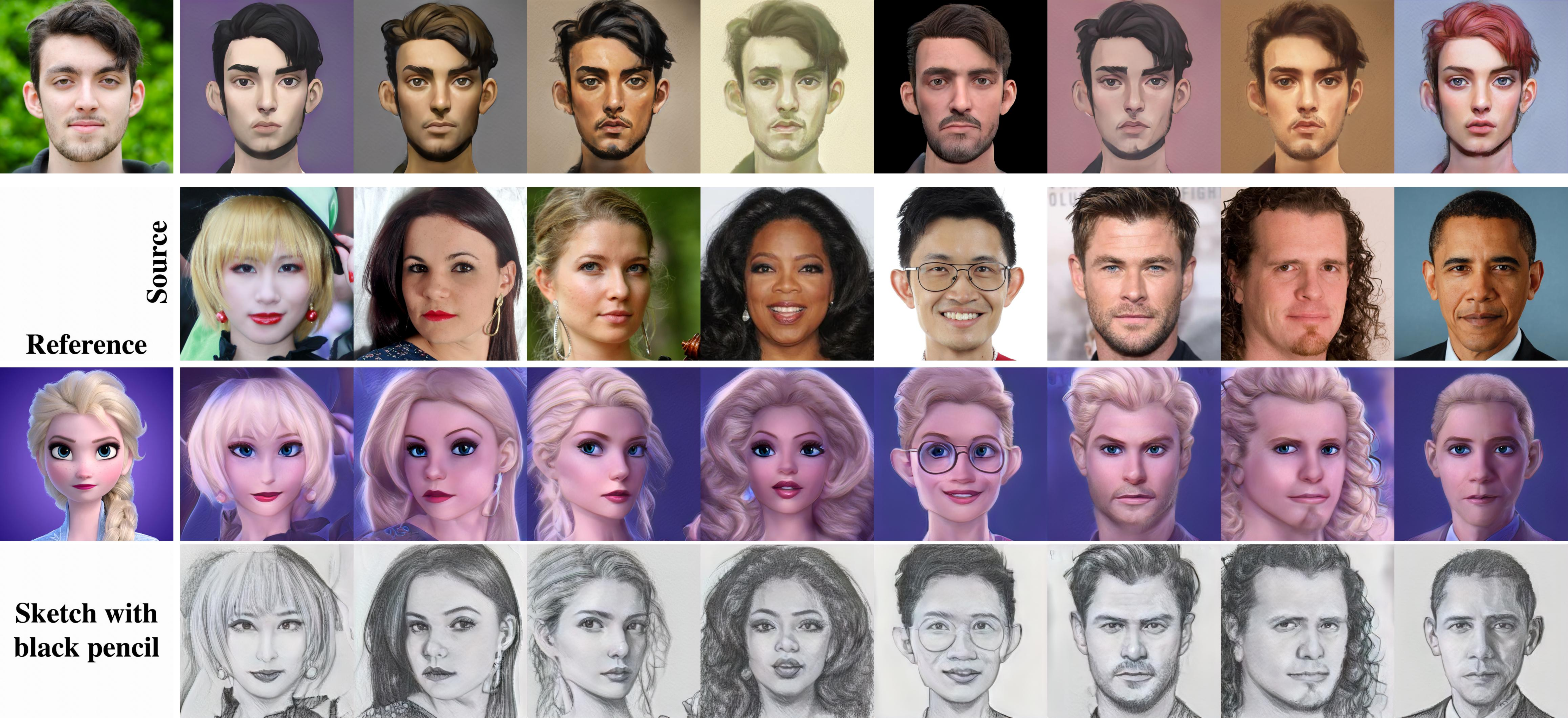}
 \centering
  \caption{Our proposed framework \shortname generates high-quality stylized faces with diverse styles (the top row) and can be applied to both one-shot and zero-shot stylization tasks (the third and fourth rows).}
\label{fig:teaser}
}

\maketitle
\begin{abstract}
In this work, we introduce a new approach for face stylization. Despite existing methods achieving impressive results in this task, there is still room for improvement in generating high-quality artistic faces with diverse styles and accurate facial reconstruction. Our proposed framework, MMFS, supports multi-modal face stylization by leveraging the strengths of StyleGAN and integrates it into an encoder-decoder architecture. Specifically, we use the mid-resolution and high-resolution layers of StyleGAN as the decoder to generate high-quality faces, while aligning its low-resolution layer with the encoder to extract and preserve input facial details. We also introduce a two-stage training strategy, where we train the encoder in the first stage to align the feature maps with StyleGAN and enable a faithful reconstruction of input faces. In the second stage, the entire network is fine-tuned with artistic data for stylized face generation. To enable the fine-tuned model to be applied in zero-shot and one-shot stylization tasks, we train an additional mapping network from the large-scale Contrastive-Language-Image-Pre-training (CLIP) space to a latent $w+$ space of fine-tuned StyleGAN. Qualitative and quantitative experiments show that our framework achieves superior performance in both one-shot and zero-shot face stylization tasks, outperforming state-of-the-art methods by a large margin.


\end{abstract}

\renewcommand{\thefootnote}{}
\footnotetext{\textsuperscript{$*$}Joint first authors.}
\footnotetext{\textsuperscript{$\dagger$}Corresponding author: jackiehuanghaibin@gmail.com}


\section{Introduction}
Artistic face stylization has been a popular research topic during the past few years, especially in the fields of computer graphics, computer vision, and machine learning~\cite{han2021exemplar,yaniv2019face,yi2019apdrawinggan}.
Given a facial image, artistic stylization methods aim to automatically transform it into a stylized version that is consistent with the input content but in a particular style. It can be used to create captivating visuals for various purposes, such as in the entertainment industry, social media, and virtual reality~\cite{abdal20233davatargan,Jamriska2018,rombach2022high}.

However, achieving high-quality face stylization is a challenging task due to the complexity of facial structure and humans' perceptual sensitivity to subtle artifacts in the output.
Compared to general image-to-image style transfer tasks, the prior knowledge humans have about facial features poses a significant challenge in preserving the original facial structure while applying the stylization process.
Moreover, various downstream application scenarios can further complicate the stylization process, including situations where one model needs to support multiple styles, only a reference style image is given, or even just a simple textual description of the desired style is provided~\cite{choi2020stargan,kwon2022clipstyler,liu2018unified}. 

In this study, we aim to address the above challenges in generating diverse stylized faces and offer a unified framework called Multi-Modal Face Stylization (\shortname). \shortname allows for the generation of diverse stylized faces and provides control over the stylization process via either a single reference image or textual description.
Our approach is based on three key observations.
First, with the advancements in generative models, it is now possible to generate realistic and diverse human faces with fine-grained details using methods such as StyleGAN~\cite{karras2019style,karras2020analyzing}. We can leverage this generative prior as a strong basis for faithful face reconstruction during stylization. 
Second, we can improve this generative prior by fine-tuning it with large artistic face datasets such as AAHQ~\cite{liu2021blendgan}, which enables us to support various styles using a single model. The architecture of StyleGAN2 also enables us to disentangle the latent space with respect to semantic attributes, enabling us to control the face content and style separately.
Finally, we can leverage the semantic power of CLIP models, which allows us to align textual descriptions with the style latent space and enables us to support cross-domain guided stylization.

In light of these observations, we propose an encoder-decoder based architecture with a two-stage training strategy for our \shortname framework. Specifically, the core of \shortname is a StyleGAN-like generator, which is trained to accurately reconstruct realistic input faces in the first stage. Unlike previous methods that use $w$ or $w+$ space for face reconstruction~\cite{abdal2019image2stylegan,karras2020analyzing}, we propose to align the low-resolution layer of StyleGAN2 with a convolution based encoder. Compared with direct projection of  the input image into a low-dimensional latent space, a convolution-based encoder preserves fine details in input facial images and results in better reconstructions.
In practice, we use a pre-trained StyleGAN2 as our generative prior and fix its weights, while training the encoder to predict the layer at the $32 \times 32$ resolution in a self-reconstruction manner. Such architecture allows us to leverage the fine-grained details of StyleGAN2 while also preserving important features of the input image.
In the second stage, we further fine-tune the entire network with artistic data for stylized face generation. The encoder is trained for style-free semantic and structural feature extraction and the decoder is trained to generate a stylized face from these structural features together with a latent style vector sampled from the latent $w$ space of StyleGAN2.
Our design also naturally enables the disentanglement of content and style which can be controlled by the encoder and the decoder separately. 

After the two-stage training, we obtain a face stylization framework that can transform a real face image into a randomly stylized version. To make the framework controllable for downstream tasks, we train an additional network to bridge the input image or textual guidance to the latent style vector. Specifically, we leverage the pre-trained CLIP model and adopt a four-layer transformer to map its space and the latent style space. Similar to the first stage, this mapping network can be trained using a self-reconstruction strategy, where random samples are generated for CLIP-$w$ paired learning.
Importantly, since the image and text are aligned in the CLIP feature space, the learned mapping network can be trained with images only but support textual guidance as well.

We conduct experiments to evaluate the effectiveness of \shortname in both one-shot and zero-shot stylization tasks. Our results demonstrate that \shortname is capable of generating high-quality stylized faces while preserving fine-grained details, as shown in~\Cref{fig:teaser}.



To summarize, our contributions are as follows:
\begin{itemize}
    \item We propose \shortname, a novel framework  which utilizes StyleGAN2 as a generative prior and integrate it into an encoder-decoder architecture. A two-stage training strategy is further designed to train \shortname for high-quality stylized face generation with diverse style support.
    
    \item \shortname is a flexible framework that can be adapted to various downstream tasks, including one-shot and zero-shot stylization. We also introduce a novel loss for better style preservation in one-shot stylizaton.
    
    

    \item \revision{Our experimental results demonstrate that \shortname generates better output than existing methods qualitatively and achieves state-of-the-art performance quantitatively.}

\end{itemize}

\section{Related Work}
\begin{figure*}
\centering
\includegraphics[width=1\linewidth]{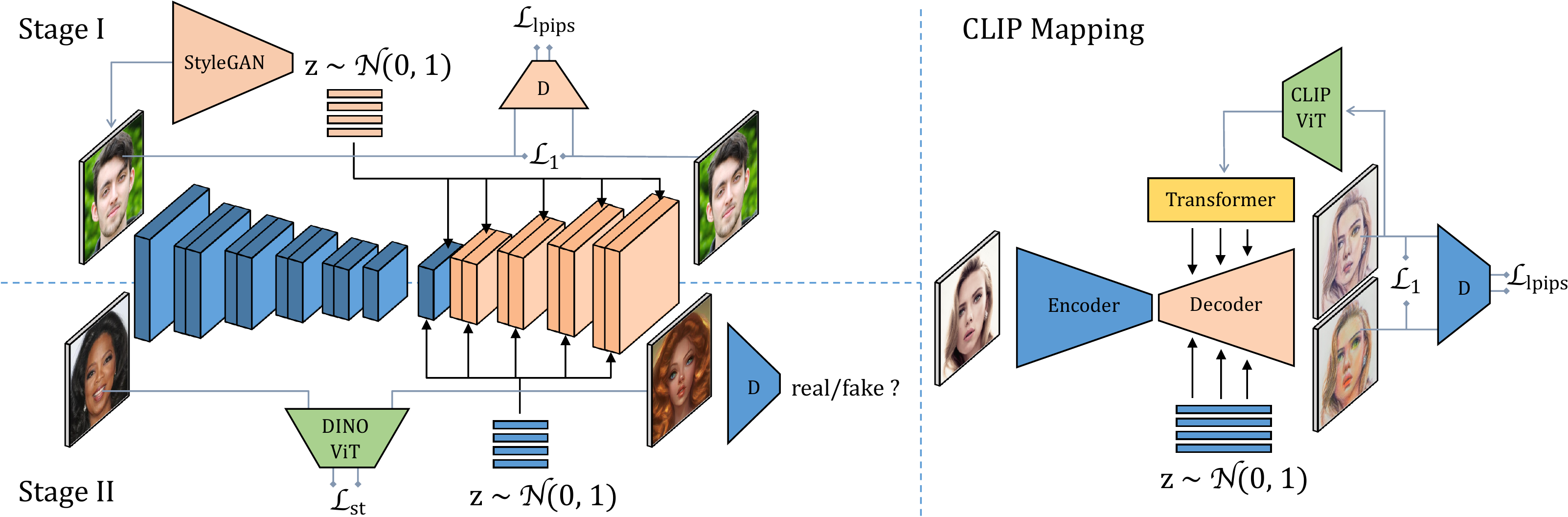}
\caption{Overview of our framework. Left: The proposed MMFS involves a two-stage training strategy, where Stage I trains an encoder to align with StyleGAN2 and to preserve fine-grained facial details, and Stage II fine-tunes the entire network for stylized face generation. Right: An additional mapping network is trained to bridge the CLIP feature space and the latent style space, providing a unified control for guided face stylization.}
\label{fig:overview}
\end{figure*}

\paragraph{Face generative model.}
Recent StyleGAN methods~\cite{karras2019style,karras2020analyzing} have substantially improved face generation benchmarks.
To accurately reconstruct and manipulate images of real faces, optimized-based GAN inversion methods~\cite{abdal2019image2stylegan,abdal2020image2stylegan++} propose several constraint functions to obtain the latent directly.
For example, II2S~\cite{zhu2020improved} introduces a $P_{N}$ space to find the trade-off point of diversity and quality.
However, inversion based methods are typically time-consuming.
As a result, a series of encoder-based approaches are widely used to achieve fidelity and editability of synthetic portraits efficiently.
pSp~\cite{richardson2021encoding} and e4e~\cite{tov2021designing} adopt a well-designed pretrained encoder network to map portrait images into the latent space.
ReStyle~\cite{alaluf2021restyle} utilizes iterative refinement to gradually modify the latent space vector via the residual-based encoder.

These inversion techniques have also been introduced for artistic face generation.
Toonify~\cite{pinkney2020resolution} uses a layer swapping scheme to interpolate between real photos and cartoon images, which preserves the texture  characteristic of the style and the structural fidelity of the photo.
StyleCariGAN~\cite{jang2021stylecarigan} achieves caricature generation using shape exaggeration blocks to modulate StyleGAN2 feature maps.
AniGAN~\cite{li2021anigan} designs special fusion blocks and a double-branch discriminator to learn domain-specific and domain-shared information.
AgileGAN~\cite{song2021agilegan} proposes a hierarchical VAE to encode the content into the $z+$ space to ensure consistency with the prior distribution.
However, these methods only transmit spatially structural signals through some stacks of one-dimensional vectors, which lead to generation results of limited fidelity.
To address this issue, we propose a spatial encoder to modulate the feature map of StyleGAN2 directly.

\paragraph{Multi-modal face stylization.}
The diversity of generation results is a key feature of face stylization models.
MUNIT~\cite{huang2018multimodal} designs two encoder-decoder modules to exchange content and style features via adaptive instance normlization~\cite{huang2017arbitrary} to allow the style latent to be sampled.
DRIT++~\cite{lee2020drit++} enhances the disentanglement of representations by introducing a domain discriminator.
StarGAN v2~\cite{choi2020stargan} allows image and latent guided generation in a style branch to enrich the diversity of control condition.
However, the decoder of these methods cannot achieve high-quality face stylization due to the lack of generative priors.
More recently, BlendGAN~\cite{liu2021blendgan} transfers the scheme in StyleGAN2 decoder and achieves desirable face stylization results in terms of both quality and variety.
Unfortunately, the decoupling ability is naturally limited by the $w/w+$ space of StyleGAN2, lacking of sufficient spatial information.
To address these weaknesses, our approach explicitly decouples the content and style information in StyleGAN2, so that the content structure is not significantly affected while changing the style.

\paragraph{Guided face stylization.}
Given out-of-domain style reference images, the model should be able to adapt to the new domain.
Some few-shot methods~\cite{li2020few,mo2020freeze,robb2020few,wang2020minegan,wang2018transferring} focus on transferring to the target domain based on several reference images.
For instance, Transferring GANs~\cite{wang2018transferring} defines the adaption of generation model by limited data as a transfer task.
FreezeD~\cite{mo2020freeze} freezes the lower layers of the discriminator to avoid overfitting. 
FSGA~\cite{ojha2021few} proposes the CDC loss to keep the diversity of generated output.
CtlGAN~\cite{wang2022ctlgan} constructs a novel encoder and the CDT loss to perform the stylization task.

Under a more strict setting of only one reference image, JoJoGAN~\cite{chong2021jojogan} perturbs the part of inversion latent to augment style.
Mind-The-Gap~\cite{zhu2022mind} constructs a geometric constraint relation in the CLIP space to shift domains while maintaining the overall structure.
Similarly, OneshotCLIP~\cite{kwon2022one} makes full use of the semantic consistency of the same portrait in the CLIP space via contrastive learning.
DiFa~\cite{zhang2022towards} designs local and global adaptation functions to enhance the diversity and fidelity.
Instead of fine-tuning the entire generator, GenDA~\cite{yang2021genda} imports two light modules called attribute adaptor and attribute classifier to refine the model.
Generalized one-shot adaptation~\cite{zhang2022generalized} focuses on preserving decorations in a new domain.
HyperNST~\cite{ruta2023hypernst} defines one-shot face stylization as a style transfer problem.

For zero-shot face stylization, the desired style information is described by input text, rather than a reference image.
StyleGAN-NADA~\cite{gal2022stylegan} uses the text direction in the CLIP space as guidance to optimize styles. 
StyleCLIP~\cite{patashnik2021styleclip} adopts a latent mapper to hierarchically adjust the latent.
TargetCLIP~\cite{chefer2022image} adds an essence vector to the source latent to make multiple views in the CLIP space.
With another usage of CLIP, CSLA~\cite{zheng2022bridging} projects the CLIP embedding into the latent space directly to edit image attributes.
To provide feasibility under different conditions, our framework is designed to support both one-shot and zero-shot face stylization simultaneously.

\section{Method}

In this section, we formally introduce \shortname. As shown in \Cref{fig:overview}, the core of the network architecture is an encoder-decoder framework integrated with StyleGAN2, which has been pre-trained with FFHQ and serves as a generative prior for high-quality face generation. To ensure stable and efficient training of this network, we propose a two-stage training strategy.
First, we train the encoder to approximate the low-resolution feature maps of the pre-trained StyleGAN2. Next, we fine-tune the entire encoder-decoder network with artistic portrait images, allowing the network to learn a domain translation from real faces to stylized ones.
To enable guided face stylization, we further train an additional mapping network from the CLIP feature space to the trained latent $w+$ space. This additional network provides a unified control for both one-shot and zero-shot face stylization. 


\subsection{Stage I: Encoder Pre-training}
\label{sec:first_stage}
As demonstrated in ~\cite{karras2019style, Kang2021GANIF}, the mid- and high-resolution layers of StyleGANs are more related to semantic attributes and details of generated faces, while the low-resolution layers are responsible for controlling facial structure.
Based on this observation, the first stage of  \shortname is to train the encoder component of the network and to align its output with the $32 \times 32$ feature maps of StyleGAN2 to ensure faithful reconstruction of the input face.



During the training process, we sample a batch of images $\referimage$ using the pre-trained StyleGAN2 generator $\generator_{sty}$ with random noises $z$.
These images and the corresponding noises are then fed into the encoder-decoder framework ($\encoder$ and $\decoder$), resulting in reconstructed images $\reconimage = \decoder(\encoder(\generator_{sty}(z)), z)$.

We use an $\mathcal{L}_1$ loss and a perceptual loss to penalize the difference between the reconstructed images and sampled images. To further preserve fine-grained details, we use the discriminator $\discriminator$ of StyleGAN2 to evaluate the perceptual loss, as suggested by JoJoGAN~\cite{chong2021jojogan}. 
Thus, the objective to optimize in this stage is:
\begin{equation}
        \lossfirststage = \loss_1(\reconimage, \referimage) \thinspace + \lambda_{perc} \sum_{l \in \{l_s\}} \loss_1(\discriminator^l(\reconimage), \discriminator^l(\referimage))
\end{equation}
where $\lambda_{perc}$ is the weight of the perceptual loss term and is set to $4.0$ empirically, $\discriminator^l(\cdot)$ denotes the $l$-th layer features extracted by the discriminator, and $\{l_s\}$ denotes the set of layers of the discriminator to compute the perceptual loss.

\revision{
We compare the reconstruction results of our encoder-decoder network in Stage I with other GAN-inversion and StyleGAN2 encoding methods, including pSp~\cite{richardson2021encoding}, e4e~\cite{tov2021designing}, and II2S~\cite{zhu2020improved}.
As shown in \Cref{fig:reconstruction}, our method can faithfully recover fine-grained details including facial features and hair regions.

\begin{figure}
    \centering
    \includegraphics[width=\linewidth]{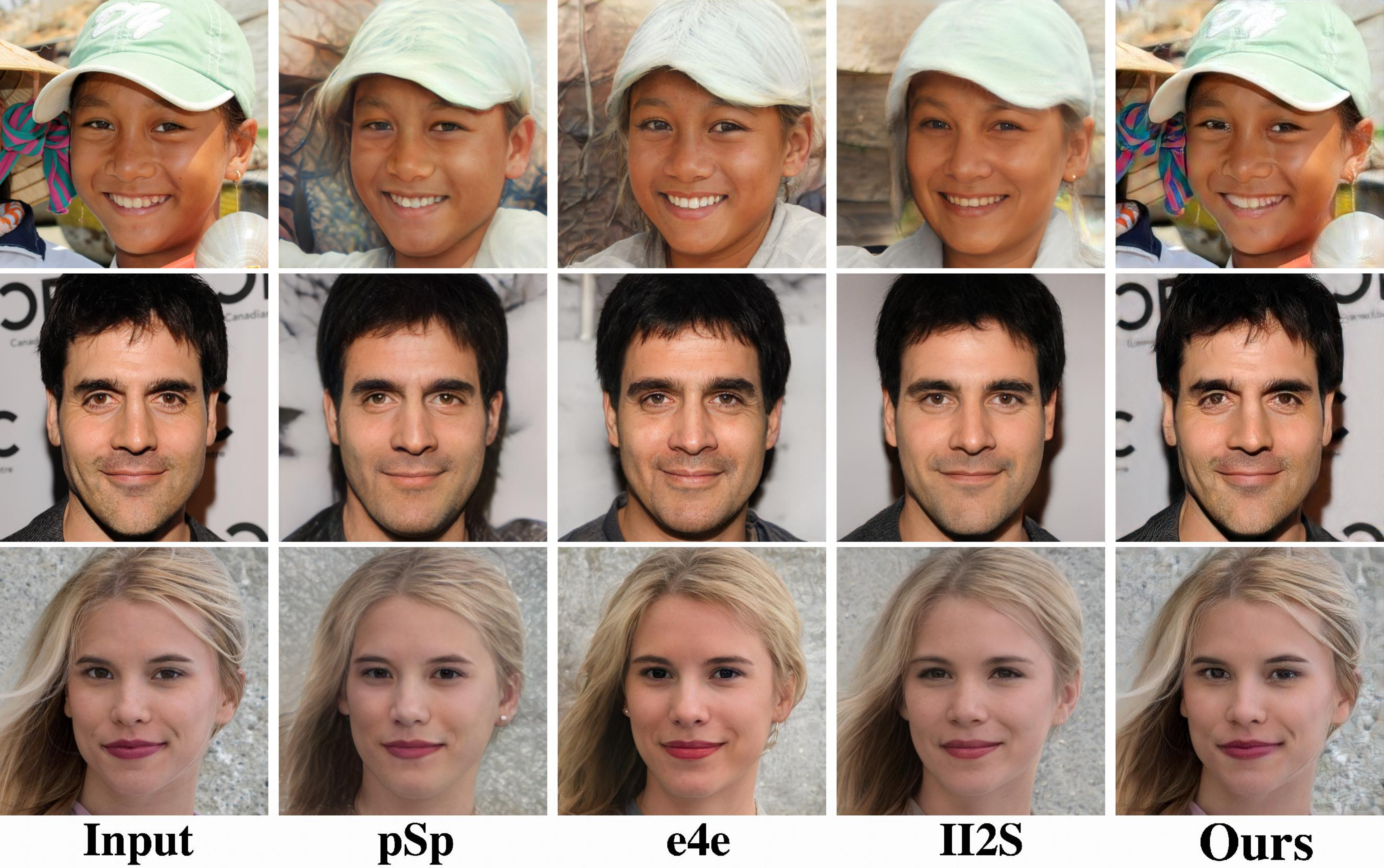}
    \caption{Comparison of reconstruction results. Our method can faithfully recover fine-grained details in local regions. }
\label{fig:reconstruction}
\end{figure}

}

\subsection{Stage II: Encoder and Decoder Fine-tuning}
\label{sec:second_stage}
In the second training stage of \shortname, we utilize the self-reconstruction model obtained in the first stage and fine-tune the entire encoder-decoder network for face stylization.
In this stage, our goal is to optimize the encoder to extract structural features that are independent of the desired artistic style, while making the decoder produce diverse stylized faces based on these structural features. 

Towards this end, we leverage the following two key components.
First, our decoder is StyleGAN-like, which can map a noise $z \in \mathcal{N}(0, 1)$ into the latent space $w$ and modulate convolution layers to produce diverse and high-fidelity results.
Here $z$ only controls the style, since the structural representation is obtained by the encoder from the input image.
To generate a stylized face image from a real face image $\inputimage$ with a noise $z$ sampled from $\mathcal{N}(0, 1)$, we use an adversarial loss to encourage our generated image to align with the distribution of stylized images $\styleimage$:
\begin{equation}
        \loss_{adv} = \mathbb{E}_{\styleimage}(\log \discriminator(\styleimage)) \thinspace + 
         \mathbb{E}_{\inputimage, z}(\log (1-\discriminator(\decoder(\encoder(\inputimage), z))))
\end{equation}
where $\discriminator(\cdot)$ denotes the prediction of discriminator.

To further constrain the structure of the generated image to be similar to the input image, we use DINO-ViT~\cite{caron2021emerging} and its self-similarity metric. Specifically, we measure the self-similarity of DINO-ViT features to obtain the structure representation~\cite{tumanyan2022splicing}:
\begin{equation}
    S^l(\mathbf{I})_{ij} = sim (K^l_i(\mathbf{I}), K^l_j(\mathbf{I}))
\end{equation}
where $sim(\cdot,\cdot)$ is cosine similarity, and $K^l_i(\cdot)$ denotes the keys of the $i$-th token in the $l$-th layer of DINO-ViT.
We measure the self-similarity for both generated image and input image, and encourage them to match each other, using the following structure loss:
\begin{equation}
    \loss_{st}=\Vert S^l(\decoder(\encoder(\inputimage), z)) - S^l(\mathbf{I_r}) \Vert_F,
\end{equation}
where $\Vert \cdot \Vert_F$ denotes the Frobenius norm.
Here we adopt the last transformer layer of DINO-ViT.

The full objective in this stage for encoder and decoder fine-tuning is:
\begin{equation}
    \losssecondstage = \loss_{adv} + \lambda_{st} \loss_{st}
\end{equation}
where $\lambda_{st}$ balances the two terms and is set to $0.5$.
We also leverage $\mathcal{R}_1$ regularization as in StyleGAN2~\cite{karras2020analyzing}. 

\subsection{Guided Face Stylization}
\label{sec:application}


After the two-stage training, we obtain a fully trained encoder-decoder network that can randomly transform a realistic face image to a stylized version.
However, this stylization process is random and beyond control.
To enable more precise and controllable stylization, we propose to train an additional mapping network that can yield a latent code in the $w+$ space from a given style image. 

\paragraph{Learning CLIP mapping.}
To enable guided stylization, we leverage a pre-trained CLIP model~\cite{radford2021learning} to extract image features and train a mapping network to convert CLIP features to $w+$ latent codes. Due to the alignment of images and text in the CLIP feature space, we can achieve text-guided stylization even if we only train on image data.

In our implementation, we use a four-layer transformer as the mapping network, with the input being the CLIP feature and a learnable positional embedding $\positionalembedding \in \mathbb{R}^{n_l \times d_c}$, where $n_l$ is the number of latent codes in the $w+$ space, and $d_c$ is the dimension of the CLIP feature.
This mapping network is trained in a self-reconstruction manner, with the same loss function as the first stage (\Cref{sec:first_stage}).
To obtain a reference image for training, we randomly stylize an input image $\inputimage$ with noise $z$: $\referimage=\decoder(\encoder(\inputimage), z)$.
We then extract the CLIP features $\clipfeature(\cdot)$ of $\referimage$, feed them into the mapping network $\mappingnetwork(\cdot)$ to obtain the corresponding latent codes, and reconstruct the image as
\begin{equation}
    \reconimage = \decoder(\encoder(\inputimage), \mappingnetwork(\clipfeature(\referimage) + \positionalembedding)).
\end{equation}
The weight of the perceptual loss is set to $\lambda_{perc}=4.0$ in this stage.
The discriminator trained in the second stage (\Cref{sec:second_stage}) is used to measure the perceptual difference.

\paragraph{Zero-shot stylization.}
Our encoder-decoder based network, which is equipped with a CLIP mapping, supports both text and image guided stylization in a single forward inference.
However, due to the limited amount of training data, the network may not always generate high-quality images that match the style prompt.
To further improve the stylization results, we introduce a fine-tuning progress that allows us to generate stylized images that are specific to a particular style prompt.
In this fine-tuning stage, we use the encoder-decoder network as a prior, while the CLIP mapping plays a role in providing an initial solution.

Given a text prompt $\textprompt$, we obtain the $w+$ latent code using CLIP mapping: $c_{w+} = \mappingnetwork(\clipfeature(\textprompt) + \positionalembedding)$, and stylize an input image $\inputimage$ with this latent code: $\styleimage=\decoder(\encoder(\inputimage), c_{w+})$.
To measure the consistency between $\styleimage$ and $\textprompt$, we use the directional CLIP loss~\cite{patashnik2021styleclip}.
Meanwhile, the structure of $\styleimage$ should be consistent with $\inputimage$.
Thus, we optimize the following objective:
\begin{equation}
    \begin{split}
        \losszeroshot(\textprompt) &= \Vert S^l(\styleimage) - S^l(\inputimage) \Vert_F \thinspace + \\
        & \lambda_{c} (1 - sim(\clipdirection(\mathbf{I_s}), \clipdirection(\textprompt)))
    \end{split}
\end{equation}
where $\lambda_{c}$ is the weight of cosine similarity, and $\clipdirection(\cdot)$ denotes the direction from the text/image anchor to CLIP features $\clipfeature(\cdot)$.
We freeze all the parameters except those of the decoder during this fine-tuning process.

\paragraph{One-shot stylization.}
For an image prompt $\imageprompt$, we perform a fine-tuning step similar to the aforementioned zero-shot stylization.
To further improve the quality of stylized image, we propose a new loss term that makes the best use of $\imageprompt$.
This new loss is built on the tokens of CLIP ViT image encoder.
Specifically, we construct a group of orthogonal basis $U \in \mathbb{R}^{d_t \times \min(d_t, n_t)}$ ($d_t$ and $n_t$ denote the dimension and number of tokens) of ViT's tokens using SVD, which constitutes a feature space that is relevant to $\imageprompt$.
If the stylized image $\styleimage$ has a style similar to the image prompt, the corresponding ViT tokens should remain unchanged as much as possible after being projected onto these orthogonal bases and then projected back.
To achieve this goal, we add the following loss term to the fine-tuning objective:
\begin{equation}
    \begin{split}
        \lossoneshot &= \losszeroshot(\imageprompt) \thinspace + \\
        & \lambda_{proj} \Vert U U^T \cliplayertoken(\styleimage) - \cliplayertoken(\styleimage) \Vert_1
    \end{split}
\end{equation}
where $\lambda_{proj}$ controls the impact of the projection loss and $\cliplayertoken(\cdot)$ denotes the $l$-th layer's tokens of the CLIP ViT image encoder.
We use the $4$th layer in all of our experiments.

\section{Experiments}
\begin{figure}
\centering
\includegraphics[width=1\linewidth]{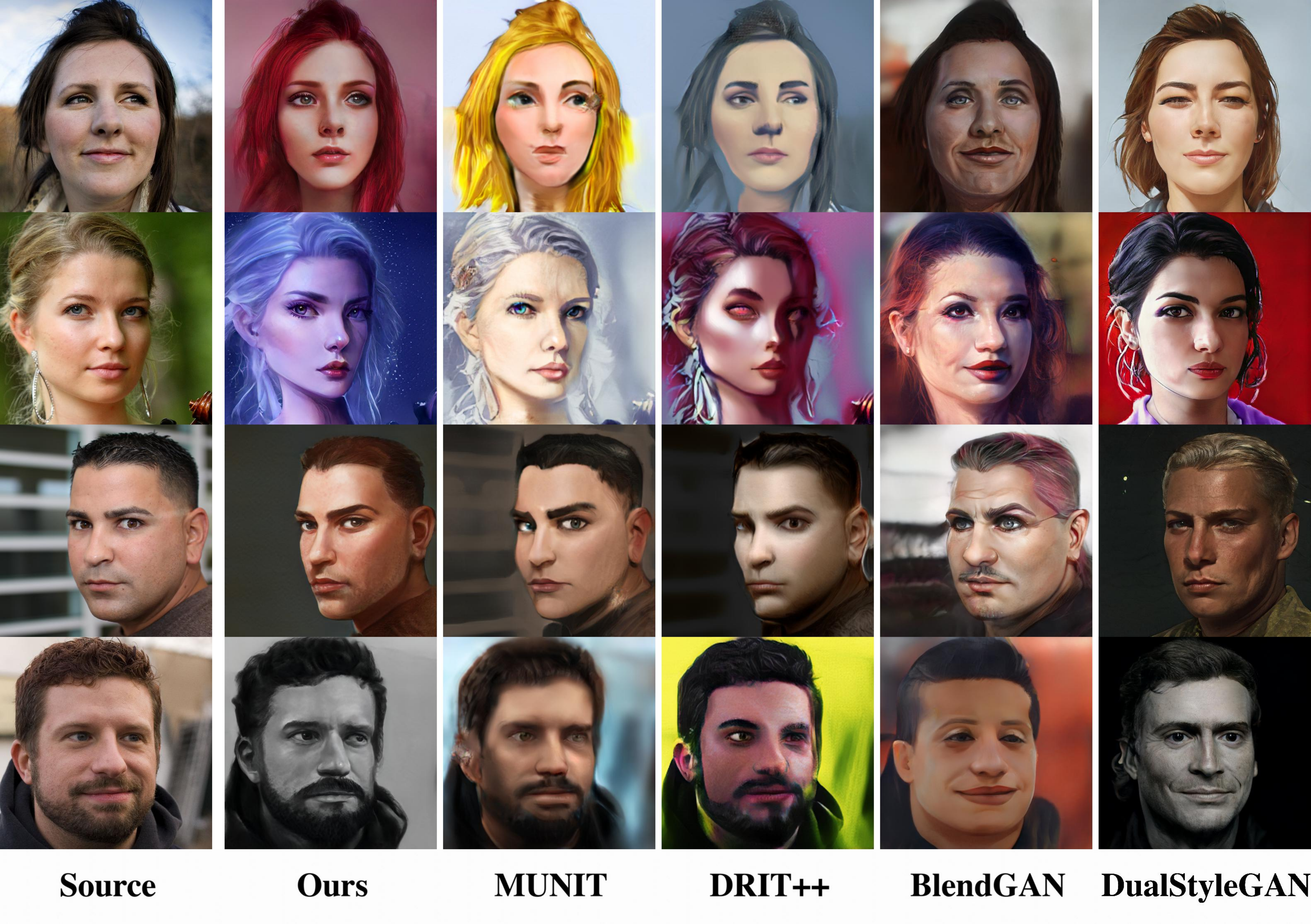} 
\caption{Qualitative comparison on random stylization. \shortname achieves both good stylization and faithful facial detail preserving.
\revision{DualStyleGAN~\cite{yang2022Pastiche} requires an input exemplar as the reference and we randomly select four images from the AAHQ dataset as exemplars for stylization.
}}
\label{fig:rand_sty}
\end{figure}

\begin{figure*}
\centering
\includegraphics[width=\linewidth]{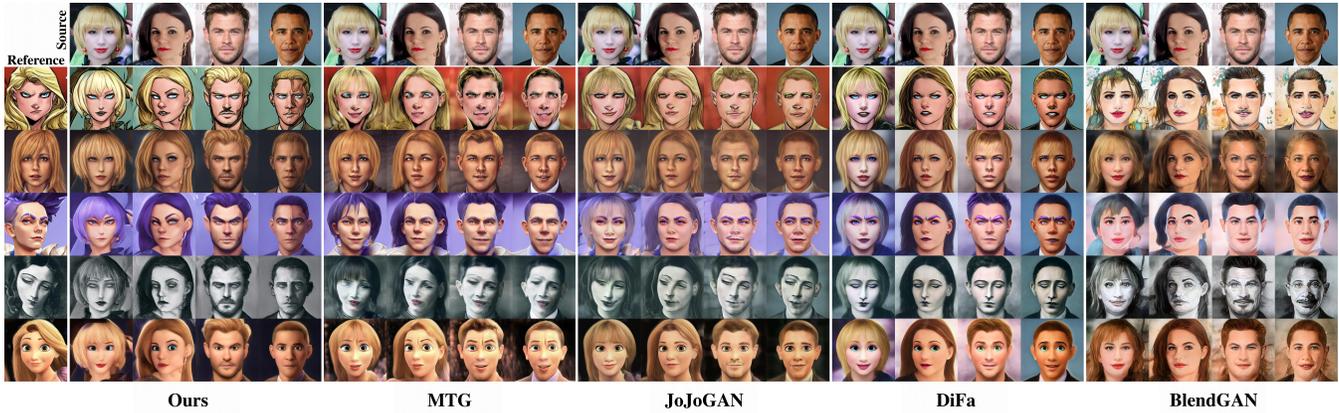}
\caption{Qualitative comparison results with several state-of-the-art methods on image-guided stylization.
}
\label{fig:qualitative_comparison_one_shot}
\end{figure*}

\begin{figure*}
\centering
\includegraphics[width=\linewidth]{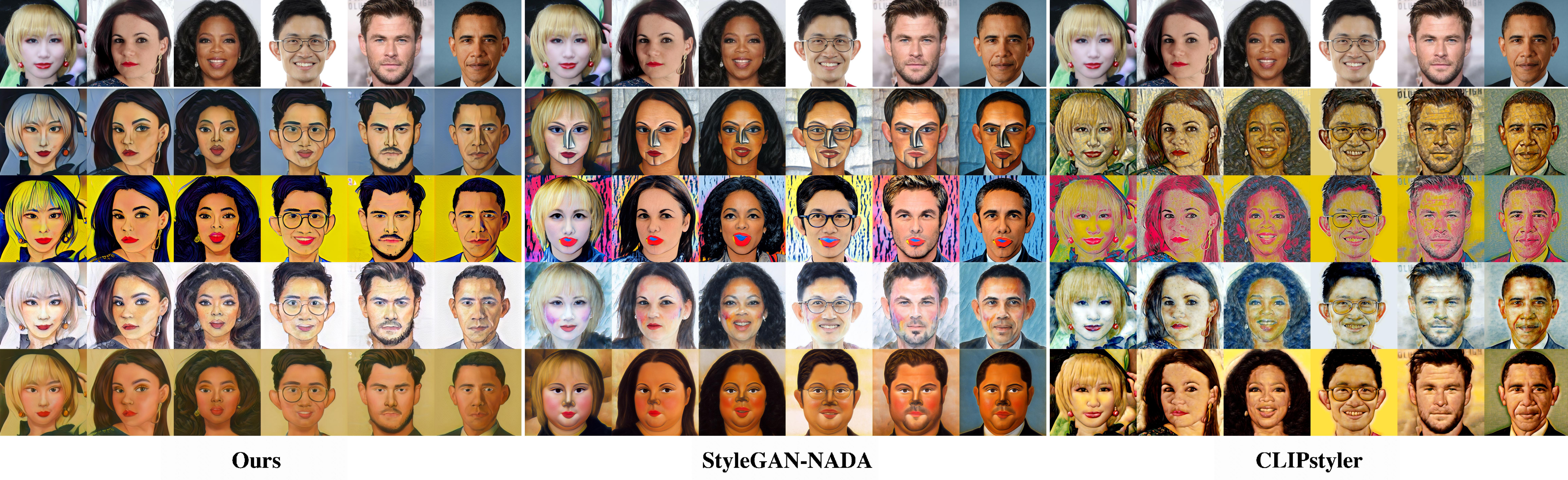}
\caption{Qualitative comparison results with several state-of-the-art methods on text-guided stylization.
The source images are shown in the first row. The text prompts for the results from the second to the fifth rows are \emph{`a cubism style painting'}, \emph{`pop art'}, \emph{`watercolor painting'}, and \emph{`painting in the style of Fernando Botero'}, respectively.
}
\label{fig:qualitative_comparison_zero_shot}
\end{figure*}

Our proposed method, \shortname, is capable of generating stylized face images with high-quality and large diversity.
In this section, we first describe our evaluation setups and then compare \shortname with several state-of-the-art baseline methods both qualitatively and quantitatively to evaluate the effectiveness of our approach in various stylization tasks.

\subsection{Experimental Setup \revision{and Implementation Details}}

\paragraph{Baseline methods.}
We compare our \shortname with several previous methods on random stylization, including MUNIT~\cite{huang2018multimodal}, DRIT++~\cite{lee2020drit++}, and BlendGAN~\cite{liu2021blendgan}.
All of them are trained on FFHQ~\cite{karras2019style} and AAHQ~\cite{liu2021blendgan} datasets.
For guided stylization, we conduct qualitative comparisons with MTG~\cite{zhu2022mind}, DiFa~\cite{zhang2022towards}, JoJoGAN~\cite{chong2021jojogan} and BlendGAN~\cite{liu2021blendgan} on one-shot stylization, and compare our method with StyleGAN-NADA~\cite{patashnik2021styleclip} and CLIPstyler~\cite{kwon2022clipstyler} on zero-shot stylization.

\paragraph{Evaluation metrics.}
We evaluate our method quantitatively from three perspectives: quality, identity-preservation, and diversity. To measure the quality, we use Frechet Inception Distance (FID)~\cite{heusel2017gans} to calculate the difference between the stylized images and samples from AAHQ. A lower FID score indicates a better stylization quality. For identity-preservation, we use the cosine similarity of Arcface~\cite{deng2019arcface} features to evaluate the variation of face identity before and after stylization. To make this metric more intuitive, we use one minus the cosine similarity (Arcface-Dist). A lower Arcface-Dist value indicates better preservation of face identity. To measure diversity, we use LPIPS~\cite{zhang2018perceptual} to calculate the perceptual difference between two randomly stylized images. A higher LPIPS value indicates larger diversity in stylization results.

\paragraph{Network training.}
We implement our method in PyTorch and train the network with Adam optimizer using a batch size of $8$.
We train the network with $10000$, $90000$, $60000$ iterations at Stage I, Stage II, and CLIP mapping, respectively.
For one-shot and zero-shot fine-tuning, we find $200$ iterations are sufficient to achieve reasonable results and avoid over-fitting. In both Stages I and II, we set the learning rate to be $0.001$, $\beta_1=0.1$ and $\beta_2=0.999$ for the optimizer. During the learning of CLIP mapping and zero/one-shot fine-tuning, the $\beta_1$ is set to $0.9$ and the learning rate is decreased to $0.0002$.
Moreover, we apply an EMA decay of $0.999$ (Stage I, Stage II, and CLIP mapping) or $0.99$ (zero/one-shot fine-tuning) for our network except for the discriminator to stabilize the training process.  

\revision{
\paragraph{Timing statistics.}
The training process takes about $30$ minutes per $1000$ iterations.
At test time, the image encoding, text encoding, random sampling, and one/zero-shot stylization step each takes $6.54$, $6.59$, $15.59$, and $15.85$ milliseconds on average, measured with single NVIDIA Tesla V100 GPU.
}


\subsection{Random Stylization}

We randomly select 10,000 face images from the FFHQ dataset, and feed them, in addition, the style code $z$ sampled from $\mathcal{N}(0, 1)$, to each method.
\Cref{tab:quant_comp} shows the quantitative comparison between our approach and previous methods.
Note that the blending indicator of BlendGAN is set to $6$ for a trade-off between stylization strength and structural consistency.
The stylized images of MUNIT and DRIT++ suffer from artifacts and blurriness while BlendGAN produces high-fidelity faces but of less noticeable style and lower diversity due to the above trade-off.
In comparison, our method reaches a better balance of consistency and stylization strength, achieving both the best visual quality and diversity, as shown in \Cref{fig:rand_sty}, thanks to the generative prior and our encoder-decoder based framework. 

\begin{table}
\caption{Quantitative comparison on random stylization.
\revision{
In each column, the best result is highlighted in \textbf{bold} while the second best number is marked with an \underline{underline}.
DualStyleGAN~\cite{yang2022Pastiche} requires an input exemplar as the reference and we randomly sample 10000 images from the AAHQ dataset to obtain the corresponding numbers.
}
}
\centering
\begin{tabular}{c|ccc}
     \toprule
     Method & FID$\downarrow$ & Arcface-Dist$\downarrow$ & LPIPS$\uparrow$ \\
     \midrule
     MUNIT~\cite{huang2018multimodal} & 18.96 & 0.511 & 0.505 \\
     DRIT++~\cite{lee2020drit++} & \underline{16.18} & 0.514 & 0.489 \\
     BlendGAN~\cite{liu2021blendgan} & 51.94 & \underline{0.509} & 0.482 \\
     \revision{DualStyleGAN~\cite{yang2022Pastiche}} & \revision{54.96} & \revision{0.582} & \revision{\textbf{0.600}} \\
     \shortname & \textbf{10.39} & \textbf{0.507} & \underline{0.583} \\
     \bottomrule
\end{tabular}
\label{tab:quant_comp}
\end{table}

\begin{figure}
\centering
\includegraphics[width=\linewidth]{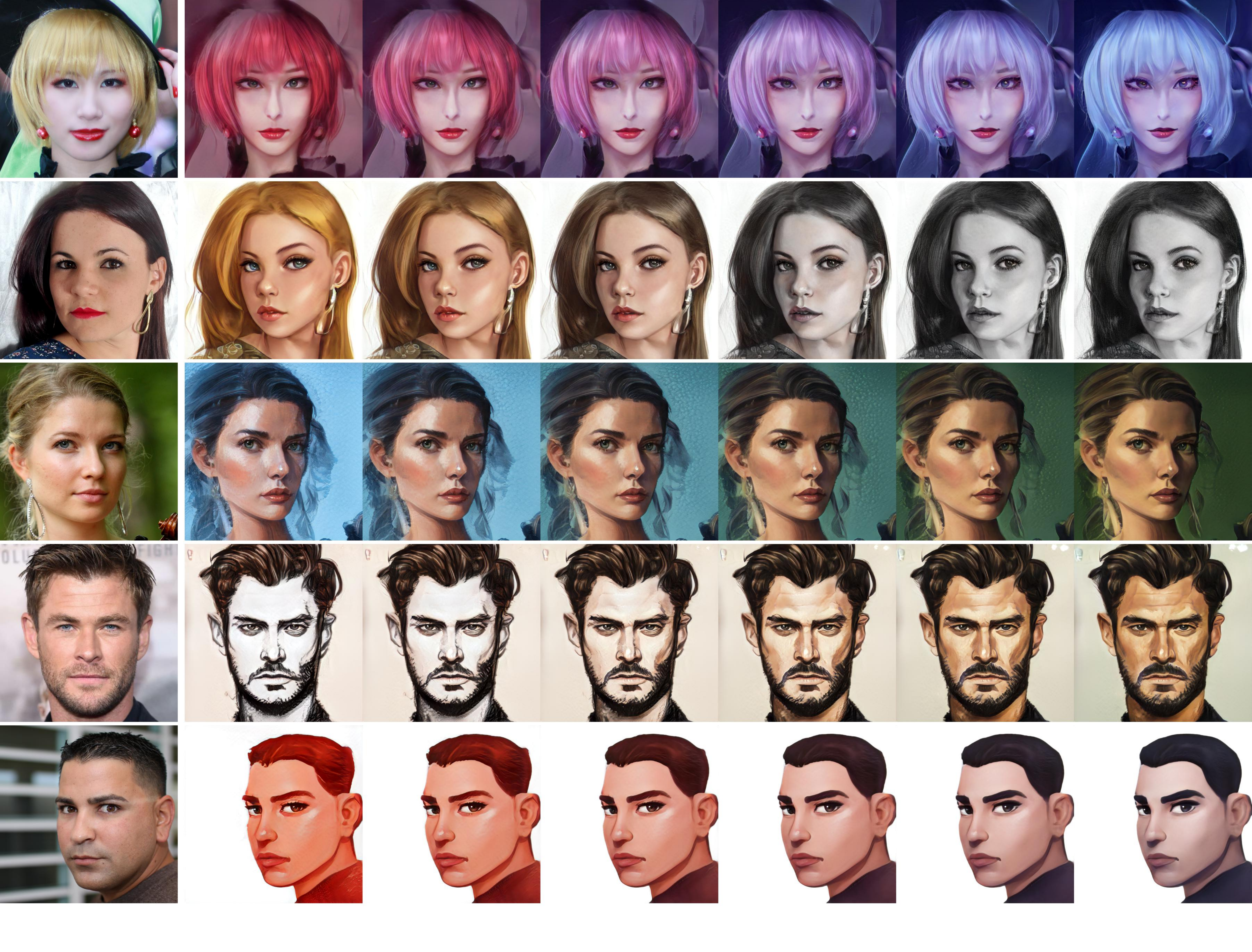}
\caption{Style interpolation results. In each row, from left to right, we present the input source image and interpolation results of two randomly sampled styles using the latent code of $w_1 \cdot \alpha + w_2 \cdot (1-\alpha)$ where $\alpha \in \{1.0, 0.8, 0.6, 0.4, 0.2, 0.0\}$.}
\label{fig:interpolation}
\end{figure}

\subsection{Guided Stylization}

\paragraph{One-shot stylization.}
In our evaluation of one-shot stylization, we compare our proposed \shortname with state-of-the-art methods, including MTG, JoJoGAN, DiFa, and BlendGAN. The results are shown in \Cref{fig:qualitative_comparison_one_shot}. From the visual comparison, we can observe that MTG and JoJoGAN tend to generate twisty facial features when given slightly irregular style images, while the eye's orientation of their stylized faces is always under the influence of reference. The overall style strength of DiFa is relatively weaker than others except for BlendGAN, and the eyes rolling back frequently occurs in its stylized faces.

Additionally, all these methods suffer from a common dilemma: they are all built on StyleGAN2, and thus a GAN inversion is indispensable to project the input image onto the StyleGAN2 $w+$ space. Unfortunately, this inversion is not perfect and leads to the loss of structure. Furthermore, some methods (e.g., MTG and BlendGAN) rely on style-mixing, which can aggravate structural changes or weaken the style.
In contrast, our proposed method integrates a StyleGAN2 into the encoder-decoder framework, allowing for both faithful reconstruction and high-quality stylization.


\begin{figure}
\centering
\includegraphics[width=1\linewidth]{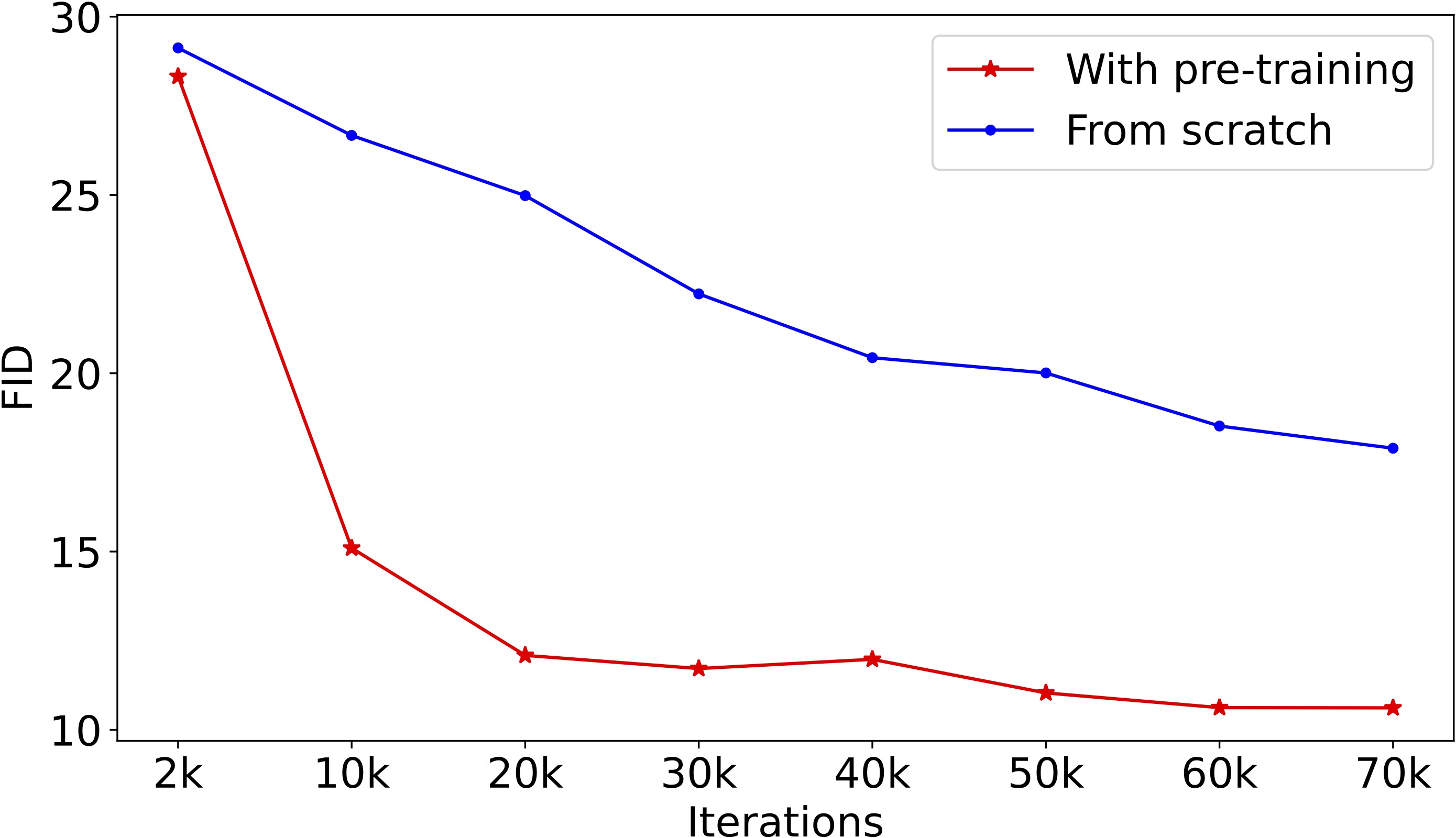}
\caption{FID curves of different training strategies. Our two-stage training leads to a significant boost in  convergence speed and much lower FID.
}
\label{fig:ab_train}
\end{figure}

\begin{figure}
\centering
\includegraphics[width=1\linewidth]{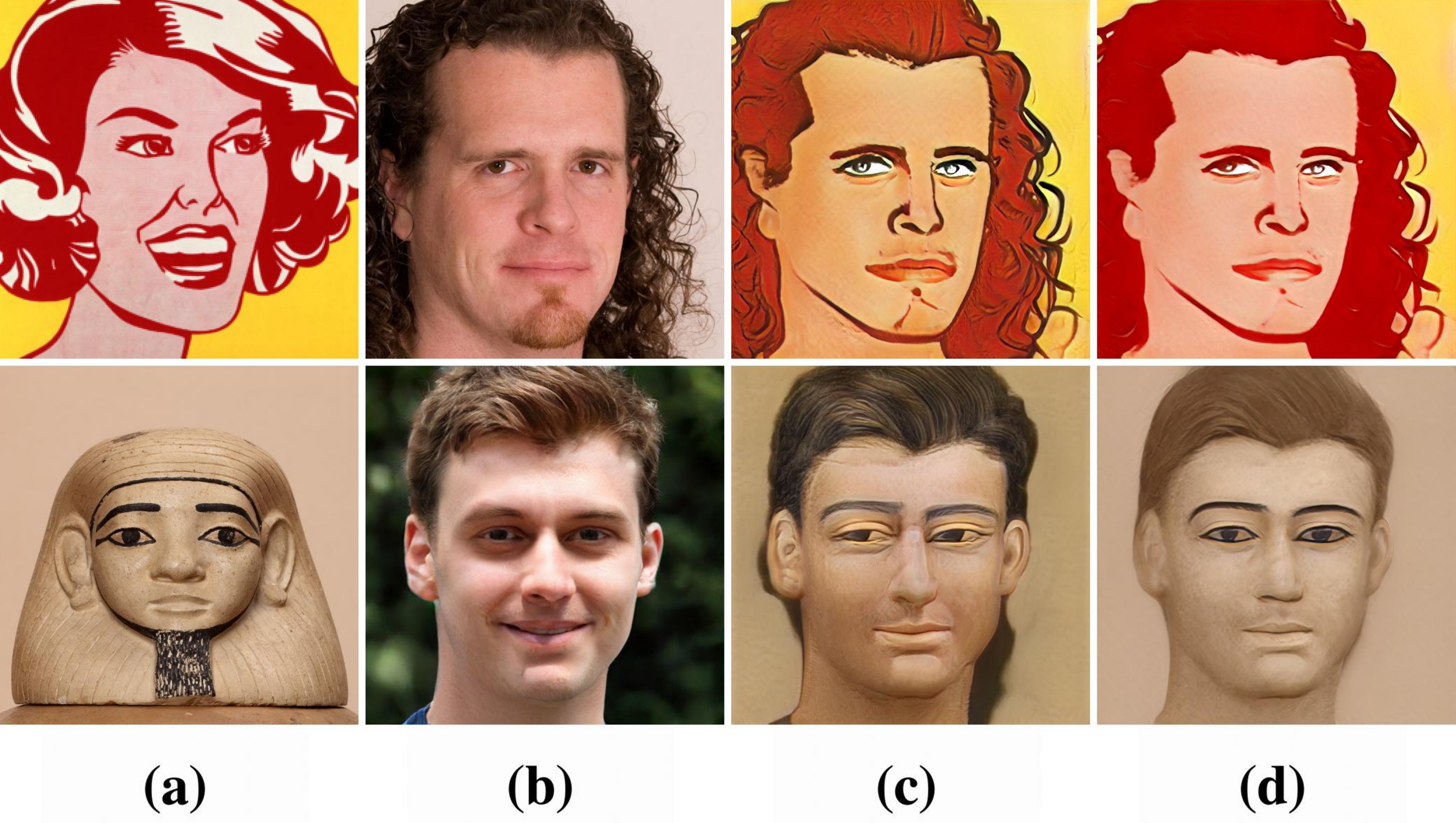}
\caption{Impact of fine-tuning and projection loss for one-shot stylization.
From left to right: (a) reference styles; (b) input source images; (c) with fine-tuning but without the projection loss, and (d) our full method, respectively.}
\label{fig:ab_one}
\end{figure}

\begin{figure}
\centering
\includegraphics[width=1\linewidth]{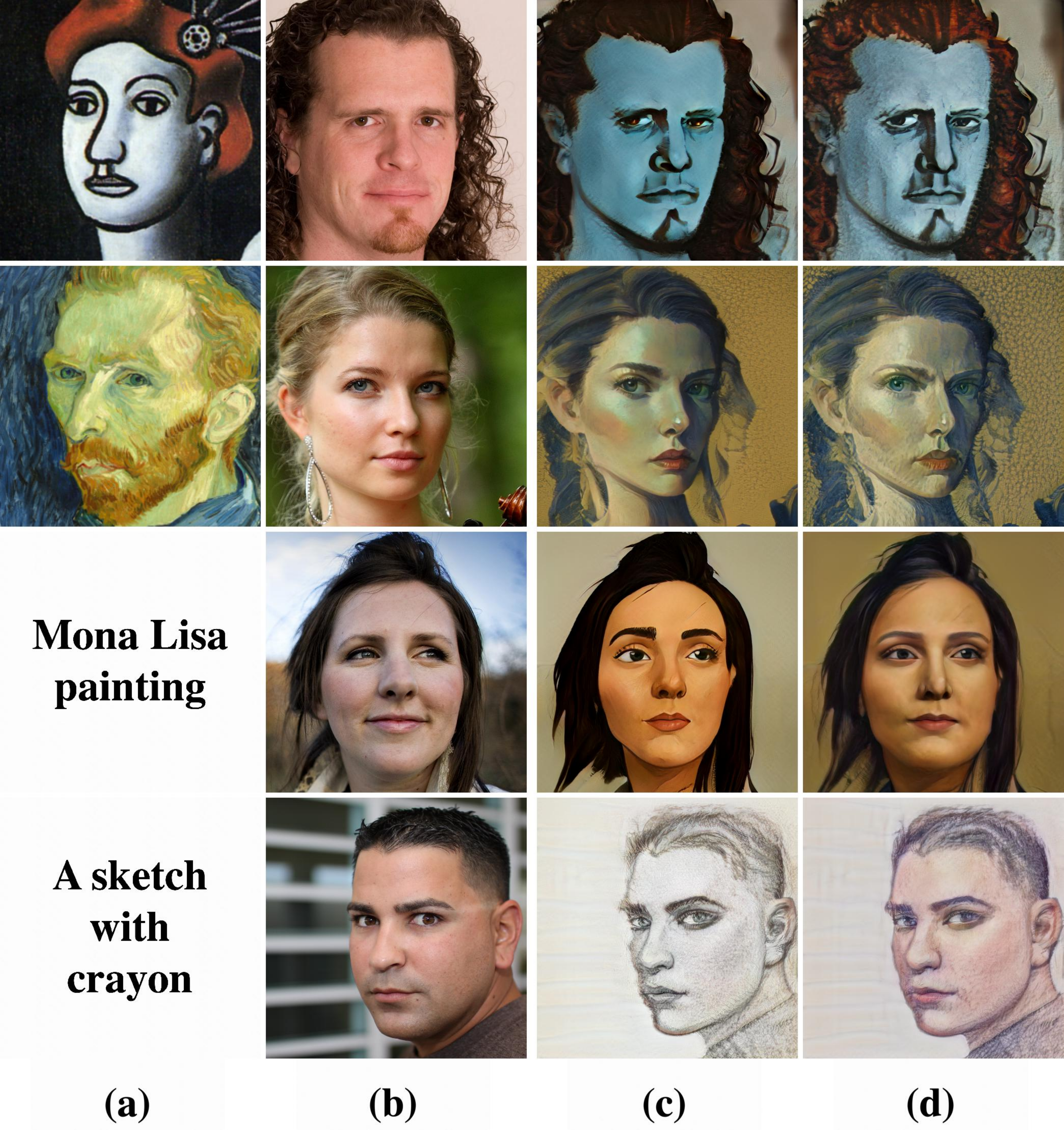}
\caption{Impact of fine-tuning for guided stylization.
From left to right: (a) reference styles (top two rows) / input text (bottom two rows); (b) input source images; (c) results based on CLIP mapping without fine-tuning and (d) fine-tuning with directional CLIP loss, respectively.}
\label{fig:ab_zero}
\end{figure}

\paragraph{Zero-shot stylization.}
In \Cref{fig:qualitative_comparison_zero_shot}, we present qualitative comparisons between our proposed \shortname and state-of-the-art methods on zero-shot stylization. Similar to StyleGAN-based methods in one-shot stylization, StyleGAN-NADA suffers from significant modifications of structure and identity in its stylized faces. Furthermore, StyleGAN-NADA often generates unwanted colorful stripes in the background region of its generated images. Meanwhile, CLIPstyler adopts an encoder-decoder framework without the need for GAN inversion, ensuring more faithful reconstruction of input faces. However, its decoder and training strategy restrict its ability to generate high-quality face images, resulting in local artifacts.
In contrast, our approach leverages a two-stage training strategy and integrates StyleGAN2 into the encoder-decoder framework, allowing for both faithful reconstruction of input faces and high-quality stylization, resulting in superior performance in zero-shot stylization.
\revision{It is also worth mentioning that recent diffusion model based methods~\cite{dhariwal2021diffusion,IDDPM} have demonstrated superior performance for zero-shot stylization. However, it remains challenging for these methods to maintain facial structure of input during the image-to-image translation process.}

\paragraph{Style interpolation.}
Thanks to the StyleGAN-like decoder, our style code in the $w+$ space can be interpolated naturally.
\Cref{fig:interpolation} shows the visual results of style interpolation, in which the style of generated images gradually changes from one to another.




\subsection{Ablation Study}

\paragraph{Training strategy.}
To evaluate the advantage of our proposed training strategy, we conduct an experiment in which we remove the generative prior and train the entire encoder-decoder from scratch in Stage II. In \Cref{fig:ab_train}, we show the FID scores with respect to the training iterations.
We observe that training from scratch leads to slower convergence and higher FID scores compared to our two-stage training strategy.
By leveraging the pre-trained StyleGAN2 and the encoder in Stage I, our fine-tuning process in Stage II converges faster and achieves consistently lower FID scores.


\paragraph{Projection loss.}
\Cref{fig:ab_one} demonstrates the positive impact of the projection loss proposed for one-shot stylization. While the directional CLIP loss encourages global semantic alignment between the stylized image and the style prompt, it may still not be sufficient enough to provide guidance for low-level features and fine-grained details. To further refine results in one-shot scenarios, our projection loss uses ViT tokens from the style reference to provide additional patch-level clues.

\paragraph{Fine-tuning.}
As shown in \Cref{fig:ab_zero}, the additional fine-tuning  for zero-shot and one-shot stylization results in stylized images that better match the style prompt. When using only the CLIP mapping, the stylized results are reasonable, but the color tone and texture details may not always be consistent with the guidance. Fine-tuning allows for the direction of the text or image prompt in the CLIP space to provide a global cue to improve these issues.



\begin{figure}
\centering
\includegraphics[width=\linewidth]{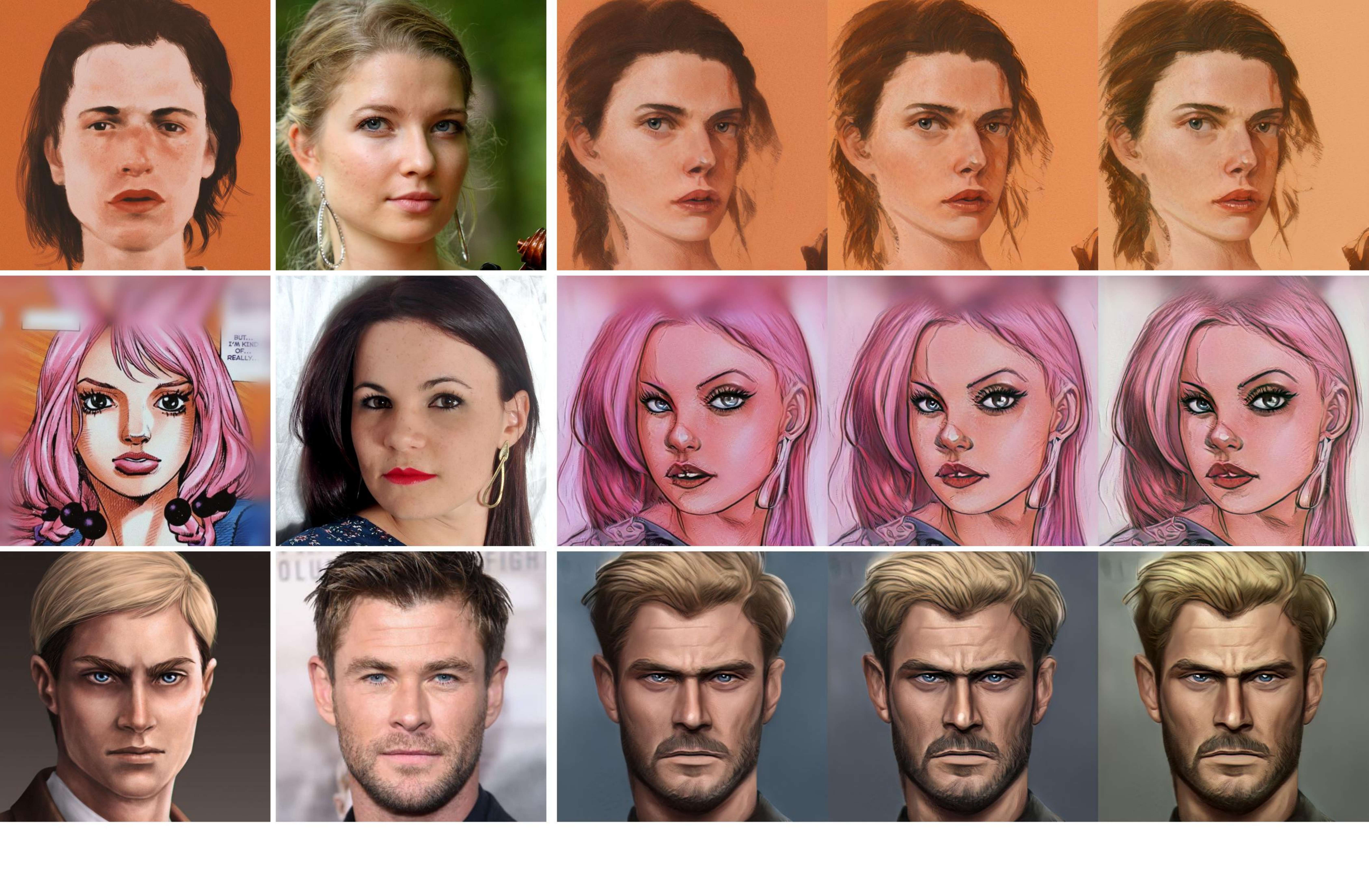}
\vspace{-2em}
\caption{Impact of different CLIP layers. From left to right: reference styles, input source images, results using the 4th, 8th, and 12th CLIP layer, respectively.}
\label{fig:clip_layer}
\end{figure}

\paragraph{CLIP features.} 
We further conduct an experiment to validate the impact of different CLIP~\cite{radford2021learning} layers in one-shot fine-tuning. As shown in \Cref{fig:clip_layer}, shallower layers contain more low-level features, leading to better color match, while deeper layers are more semantic. We applied the proposed projection loss on deep layers which provides similar guidance as directional CLIP loss. We currently choose the 4th layer of CLIP features for a trade-off between low-level and semantic features, which works well in our experiments.

\begin{figure}
\centering
\includegraphics[width=\linewidth]{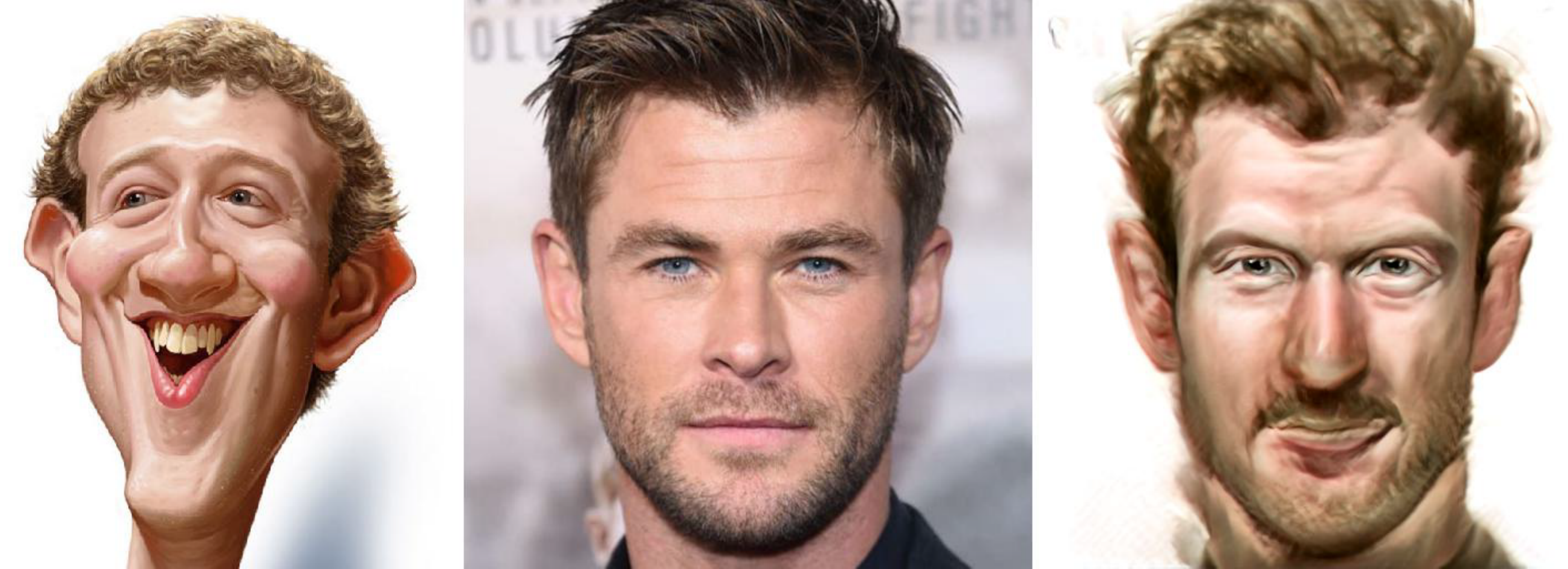}
\caption{ Our method has limitations in handling significant geometric deformations. Given a reference style such as a caricature (on the left), our method is unable to deform the facial structure accordingly (on the right) and instead transfers the source input with a similar texture appearance.}
\label{fig:limitations}
\end{figure}

\begin{figure}
\centering
\includegraphics[width=\linewidth]{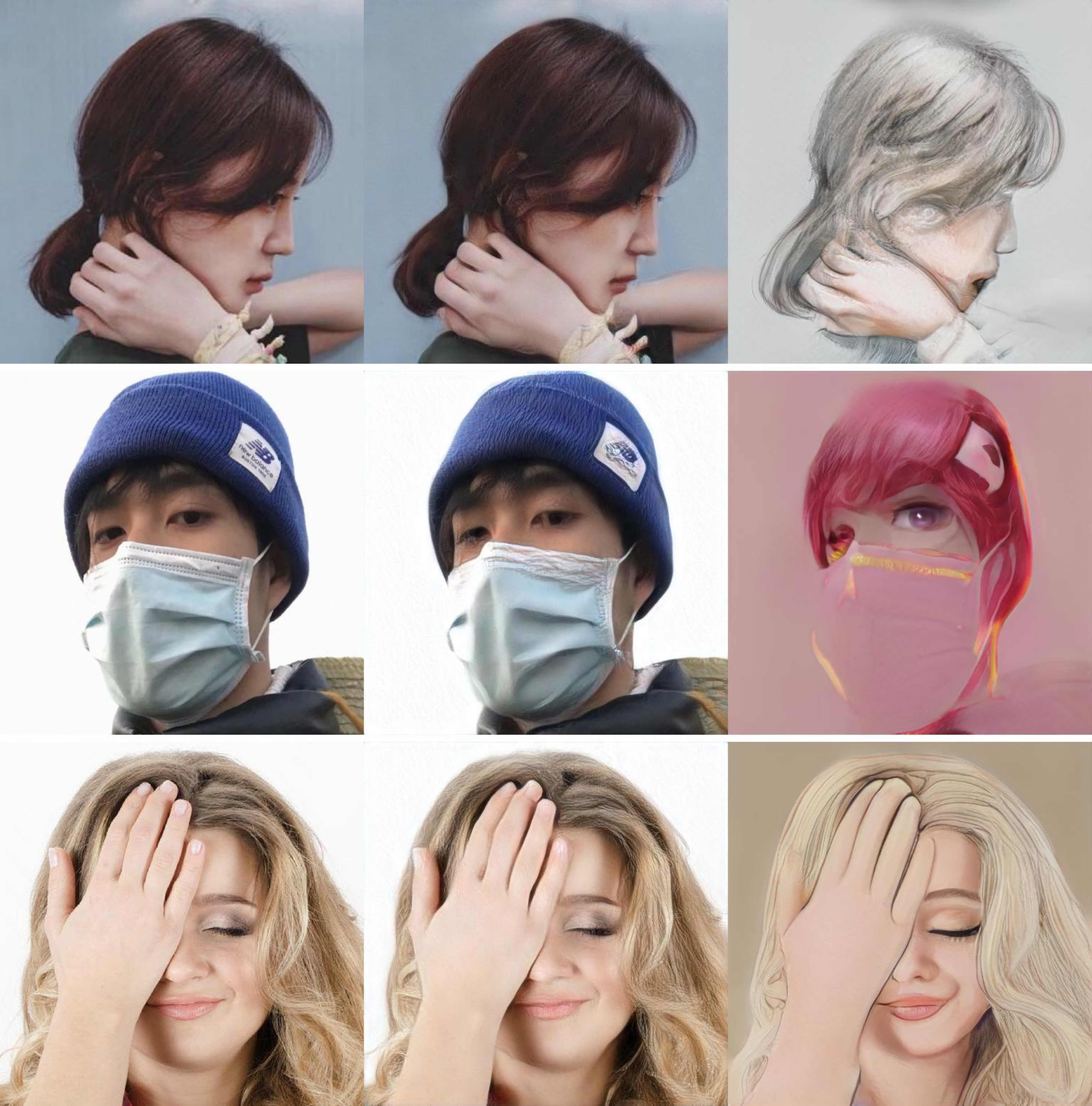}
\caption{Out-of-distribution results. From left to right: input images, reconstruction results, and stylized output.}
\label{fig:ood}
\end{figure}

\subsection{Limitations}


Although \shortname achieves high-quality face stylization with large diversity in all the cases of random/one-shot/zero-shot stylization, there is still potential for further improvement. Specifically, our current implementation does not support significant geometric deformation such as caricature, as shown in \Cref{fig:limitations}. This issue may be addressed or alleviated by incorporating a learnable deformation module into the network.
Moreover, our generated images are limited within the cropped region of FFHQ.
\revision{Our method also has limited performance for input with large pose variations or significant occlusions that are out-of-distribution of FFHQ, as shown in Figure \ref{fig:ood}.}
Integrating our \shortname with VToonify~\cite{yang2022vtoonify} is a promising future direction to unlock the power for full-size image stylization.

\section{Conclusion}
In this study, we present \shortname as a novel framework that leverages the strengths of StyleGAN2 and an encoder-decoder architecture to generate high-quality and diverse stylized faces while preserving fine-grained details. With our two-stage training strategy and CLIP-guided mapping network, MMFS offers a flexible and effective approach for one-shot and zero-shot stylization tasks. Our experimental results demonstrate that MMFS outperforms state-of-the-art methods by a large margin in both quantitative and qualitative evaluations. Overall, MMFS presents a promising solution for artistic face stylization with a wide range of potential applications. It is also worth exploring to expand this generative prior based framework to other applications beyond face stylization.



\paragraph{Acknowledgements.}
We thank the anonymous reviewers for their valuable feedback and constructive suggestions.

\bibliographystyle{eg-alpha-doi}
\bibliography{egbib}



\end{document}


\title{Multi-Modal Face Stylization with a Generative Prior -- Supplementary Materials}

\author{First Author\\
Institution1\\
Institution1 address\\
{\tt\small firstauthor@i1.org}
\and
Second Author\\
Institution2\\
First line of institution2 address\\
{\tt\small secondauthor@i2.org}
}

\maketitle

\section{Implementation Details}

\begin{table}[h]
\caption{Network architecture.}
\centering
\begin{tabular}{c|cc}
     \toprule
     & Layer & Channel \\
     \midrule
     \multirow{6}*{Encoder} & Conv & 64 \\
     ~ & ResBlock & 64 \\
     ~ & ResBlock & 128 \\
     ~ & ResBlock & 256 \\
     ~ & ResBlock & 512 \\
     ~ & Conv & 512 \\
     \midrule
     \multirow{5}*{Decoder} & StyledConv + ToRGB & 512 \\
     ~ & $2\times$StyledConv + ToRGB & 512 \\
     ~ & $2\times$StyledConv + ToRGB & 256 \\
     ~ & $2\times$StyledConv + ToRGB & 128 \\
     ~ & $2\times$StyledConv + ToRGB & 64 \\
     \bottomrule
\end{tabular}
\label{tab:net_arch}
\end{table}

\Cref{tab:net_arch} details the network architecture of the proposed encoder-decoder framework, where \emph{`Conv'}, \emph{`ResBlock'}, \emph{`StyledConv'} and \emph{`ToRGB'} use the same structures as in StyleGAN2~\cite{karras2020analyzing} and are developed based on a PyTorch implementation\footnote{\url{https://github.com/rosinality/stylegan2-pytorch}}.

Our method is trained on a single NVIDIA Tesla V100 with a batch size of $8$.
The training process takes about $30$ minutes per $1000$ iterations.
We train the network with $10000$, $90000$, $60000$ iterations at Stage I, Stage II, and CLIP mapping, respectively.
For one-shot and zero-shot fine-tuning, we find $200$ iterations are sufficient to achieve reasonable results and avoid over-fitting.

\section{Additional Results}

\begin{figure}
\centering
\includegraphics[width=1\linewidth]{images/supp/fid-curve-all.pdf}
\vspace{-2em}
\caption{FID curves of different $\lambda_{st}$ values.
}
\label{fig:ab_fid_loss_st}
\vspace{-1em}
\end{figure}

\begin{figure}
\centering
\includegraphics[width=1\linewidth]{images/supp/arc-curve-all.pdf}
\vspace{-2em}
\caption{Arcface-Dist curves of different $\lambda_{st}$ values.
}
\label{fig:ab_arc_loss_st}
\vspace{-1em}
\end{figure}

\Cref{fig:ab_fid_loss_st,fig:ab_arc_loss_st} illustrate the impact of the structure loss on FID~\cite{heusel2017gans} and Arcface-Dist~\cite{deng2019arcface} w.r.t different weights of $\lambda_{st}$.
As we can see, a larger $\lambda_{st}$ preserves the identity better but leads to lower stylization quality.
The curves of Arcface-Dist go increasingly due to the pre-training in self-reconstruction manner at Stage I, and the FID curves oscillate more heavily given a larger $\lambda_{st}$, indicating more unstable training.
Thus, we select the model trained with $\lambda_{st}=0.5$ after $90000$ iterations for a trade-off between quality and identity-preservation.
\begin{figure*}
\centering
\includegraphics[width=0.9\linewidth]{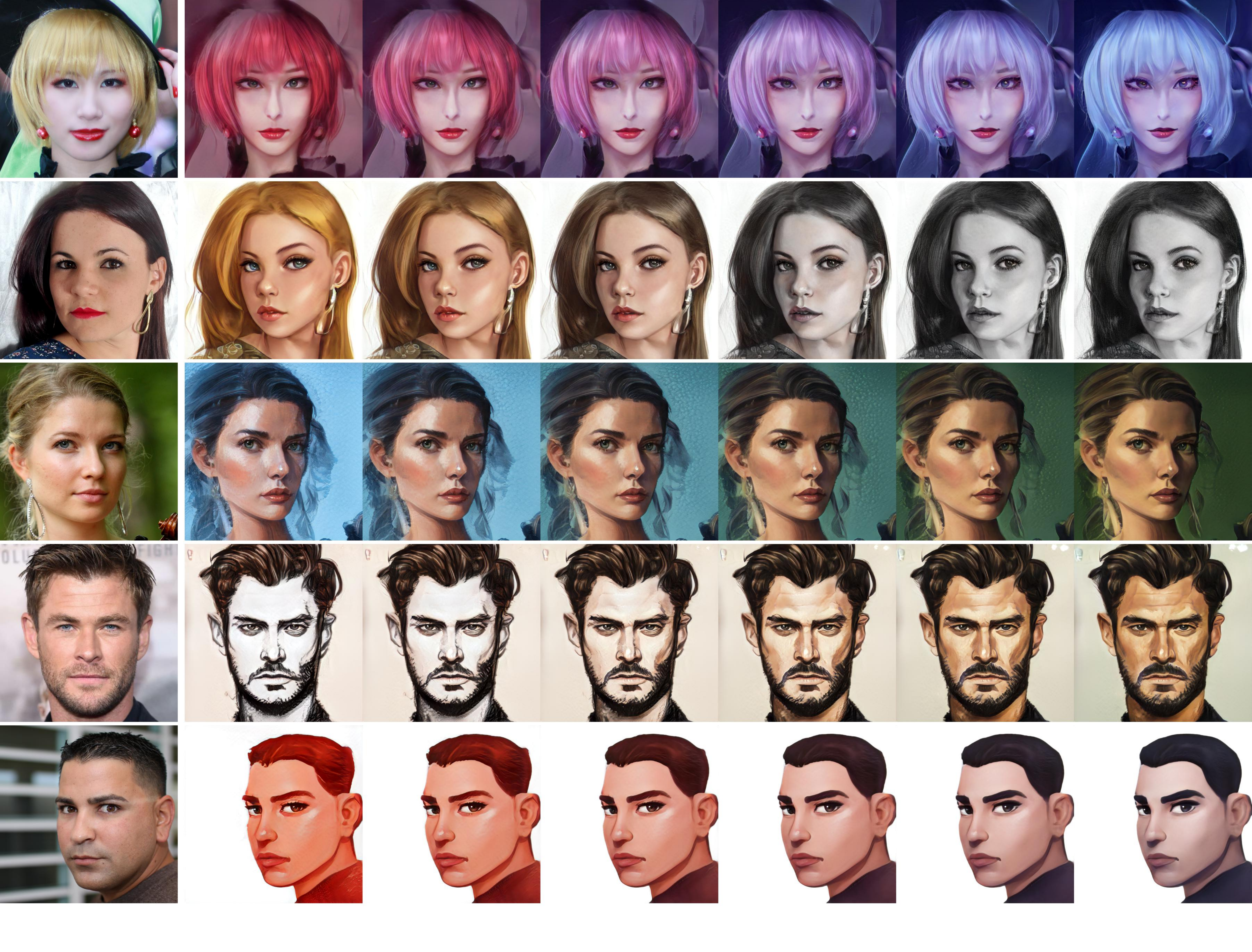}
\vspace{-2em}
\caption{Style interpolation results. In each row, from left to right, we present the input source image and interpolation results of two randomly sampled styles using the latent code of $w_1 * \alpha + w_2 * (1-\alpha)$ where $\alpha \in \{1.0, 0.8, 0.6, 0.4, 0.2, 0.0\}$.}
\label{fig:interpolation}
\end{figure*}

\begin{figure}
\centering
\includegraphics[width=\linewidth]{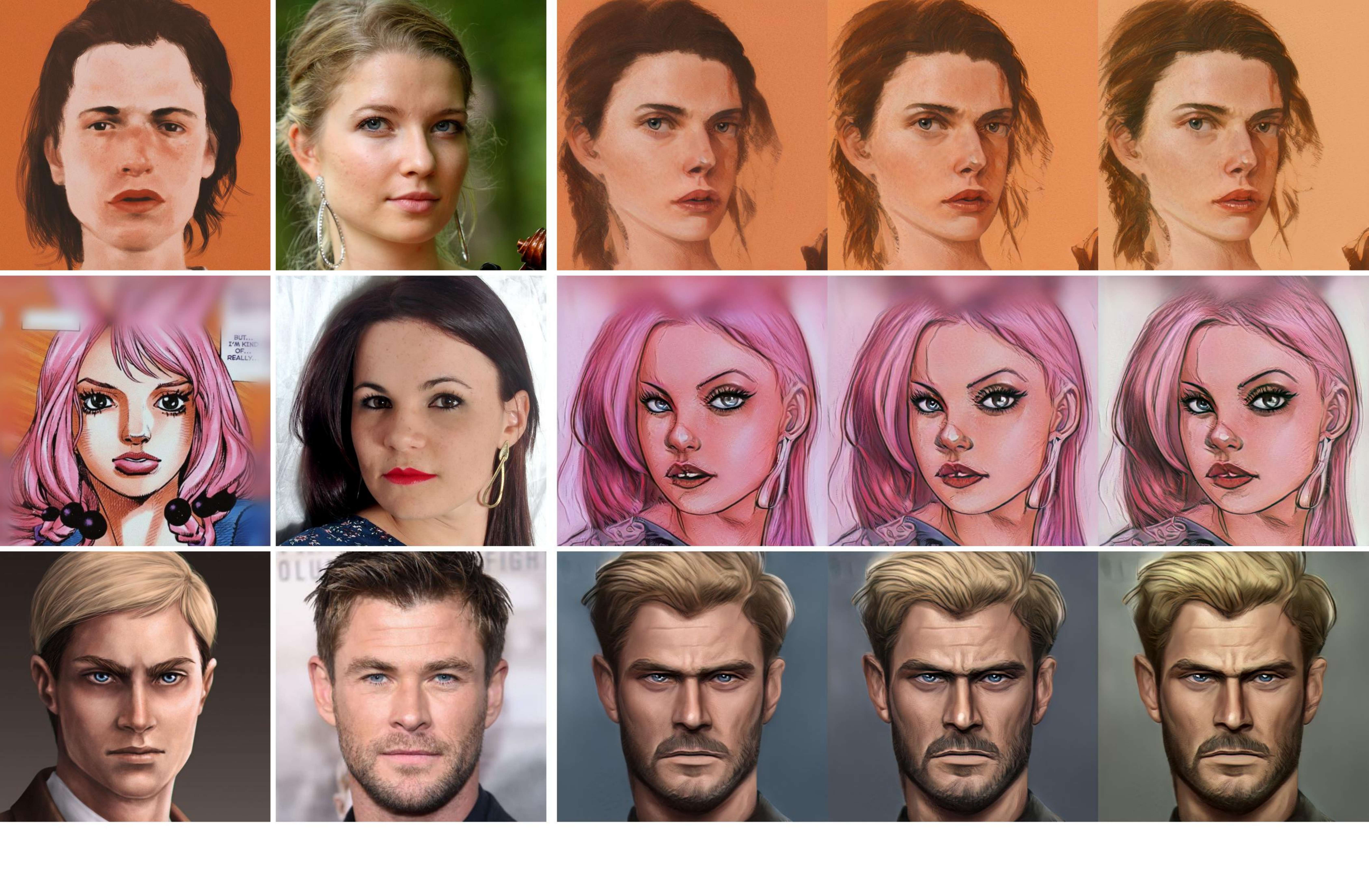}
\vspace{-2em}
\caption{Impact of different CLIP layers. From left to right: reference styles, input source images, results using the 4th, 8th, and 12th CLIP layer, respectively.}
\label{fig:clip_layer}
\end{figure}
Thanks to the StyleGAN-like decoder, our style code in the $w+$ space can be interpolated naturally.
\Cref{fig:interpolation} shows the visual results of style interpolation, in which the style of generated images gradually changes from one to another.

We further conduct an experiment to validate the impact of different CLIP~\cite{radford2021learning} layers in one-shot fine-tuning. As shown in \Cref{fig:clip_layer}, shallower layers contain more low-level features, leading to better color match, while deeper layers are more semantic. We applied the proposed projection loss on deep layers which provides similar guidance as directional CLIP loss.
We currently choose the 4th layer of CLIP features for a trade-off between low-level and semantic features, which works well in our experiments.


{\small
\bibliographystyle{ieee_fullname}
\bibliography{egbib}
}